\documentclass[letterpaper, submit]{AAS}			

\usepackage{bm}
\usepackage{amsmath}
\usepackage{amssymb}
\usepackage{amsfonts}
\usepackage{subfigure}
\usepackage[colorlinks=true, pdfstartview=FitV, linkcolor=black, citecolor= black, urlcolor= black]{hyperref}
\usepackage{overcite}
\usepackage{footnpag}			      	

\usepackage{algorithm}
\usepackage{algorithmicx}
\usepackage{algpseudocode}

\algnewcommand{\Initialize}[1]{%
  \State \textbf{Initialize:}
  \State \hspace*{\algorithmicindent}\parbox[t]{0.8\linewidth}{\raggedright #1}
}
\usepackage{subfigmat}
\usepackage{comment}
\usepackage{caption}

\makeatletter
\newif\ifcap@minipage
\cap@minipagefalse
\let\minipagebak\minipage
\def\minipage{\cap@minipagetrue\minipagebak}
\let\@captionbak\@caption
\def\@caption{%
 \ifcap@minipage
  \capwidth\hsize\ecapwidth\hsize
 \fi
 \@captionbak}
\makeatother

\begin{document}

\title{DEEP REINFORCEMENT LEARNING FOR SAFE LANDING SITE SELECTION WITH CONCURRENT CONSIDERATION OF DIVERT MANEUVERS \footnote{This paper is an updated version of Paper AAS 20-583 presented at the AAS/AIAA Astrodynamics Specialist Conference, Online, in 2020.}}

\author{Keidai Iiyama\thanks{Master Student, Department of Aeronautics and Astronautics, The University of Tokyo, 7-3-1, Hongo,Bunkyo-ku,Tokyo, Japan.},  
Kento Tomita\thanks{Ph.D. Student, School of Aerospace Engineering, Georgia Institute of Technology, Atlanta, GA, 30332.},
Bhavi A. Jagatia\thanks{Master Student, School of Aerospace Engineering, Georgia Institute of Technology, Atlanta, GA, 30332.},
Tatsuwaki Nakagawa\thanks{Bachelor Student, School of Aerospace Engineering, Georgia Institute of Technology, Atlanta, GA, 30332.},
\ and Koki Ho\thanks{Assistant Professor, School of Aerospace Engineering, Georgia Institute of Technology, Atlanta, GA, 30332.}
}

\maketitle{}

\begin{abstract}
This research proposes a new integrated framework for identifying safe landing locations and planning in-flight divert maneuvers. The state-of-the-art algorithms for landing zone selection utilize local terrain features such as slopes and roughness to judge the safety and priority of the landing point. However, when there are additional chances of observation and diverting in the future, these algorithms are not able to evaluate the safety of the decision itself to target the selected landing point considering the overall descent trajectory. In response to this challenge, we propose a reinforcement learning framework that optimizes a landing site selection strategy concurrently with a guidance and control strategy to the target landing site. The trained agent could evaluate and select landing sites with explicit consideration of the terrain features, quality of future observations, and control to achieve a safe and efficient landing trajectory at a system-level. The proposed framework was able to achieve 94.8 $\%$ of successful landing in highly challenging landing sites where over 80$\%$ of the area around the initial target lading point is hazardous, by effectively updating the target landing site and feedback control gain during descent.
\end{abstract}

\section{Introduction}
\begin{figure}[htbp]
 \begin{minipage}{0.45\hsize}
  \centering
    \includegraphics[height=55mm]{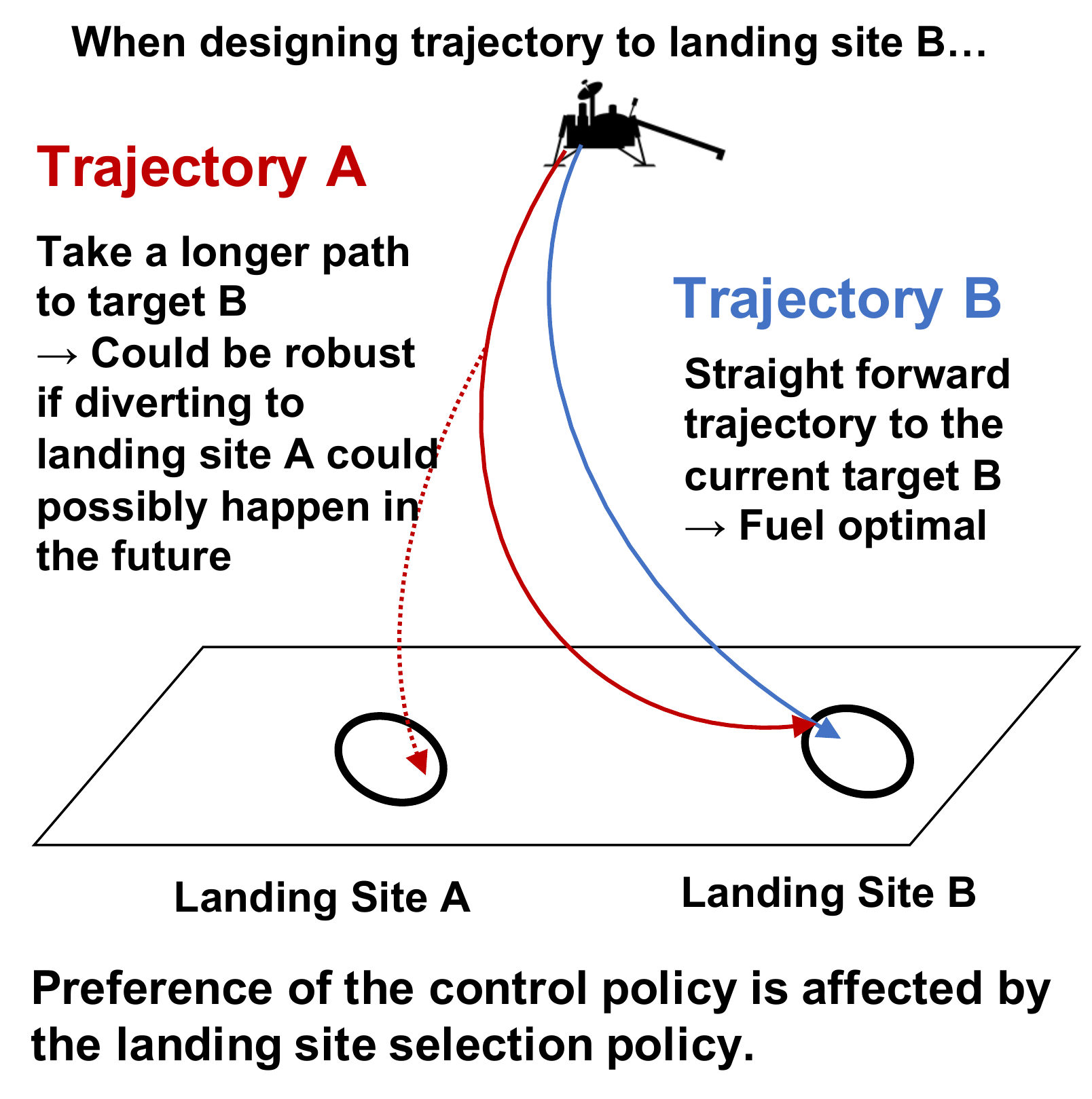}
    \caption{Landing site selection and control}
    \label{fig:coupling_ls_control}
 \end{minipage}
 \begin{minipage}{0.55\hsize}
  \centering
   \includegraphics[height=55mm]{Figures/concept_observability.pdf}
   \caption{Observability and trajectory}
  \label{fig:coupling_observability}
 \end{minipage}
\end{figure}

On-board hazard detection and avoidance (HDA) capabilities are essential to enable new mission concepts that involve planetary surface operations. With a quick assessment of the perceived terrain data (e.g., DEM, visible spectrum map, or a combination thereof) from optical and/or LIDAR sensors, the HDA technology creates a map of probability of safety for prioritizing candidate landing zones. 
NASA has been actively developing the technology for Precise landing and Hazard Avoidance (PL\& HA) \cite{prop_8, prop_9,prop_10,prop_11,prop_12,prop_13}, and Safe and Precise Landing Integrated Capabilities Evolution (SPLICE) project is currently underway to develop next generation PL\&HA technologies \cite{Restrepo2019, Cianciolo2019, Carson2019}. 
The HDA algorithm being developed for SPLICE directly leverages the HDA algorithms from the previous Autonomous Landing Hazard Avoidance Technology (ALHAT) project. ALHAT is capable of quickly assessing the DEM on-board and in real time during the descent, and assigns a probability of a safe landing to each pixel on the map. \cite{prop_2,prop_3, prop_14, prop_15, prop_16, prop_17, prop_18, prop_19, prop_20, prop_21, prop_22}. ALHAT's HDA capability along with the terrain relative navigation and hazard detection functions was successfully demonstrated during the hardware-in-the-loop testing on the Morpheus Vertical Testbed \cite{prop_18, prop_23}. However, a critical caveat of ALHAT is that it only selects a list of discrete landing points based on local static information (e.g., slopes, roughness, etc.). To consider the feasibility of the divert maneuver, a landing selection algorithm that calculates approximate landing footprint has been presented \cite{Serrano2006, Ploen2009}. To reflect predetermined scientific values of each landing site in the landing site selection process, landing site selection methods that leverages Bayesian networks has been proposed \cite{prop_13}. Cui, et al. (2017) proposed a method that calculates synthetic landing area assessment criteria based on terrain safety, fuel consumption, and touchdown performance during descent \cite{Cui2017}.

In these previously proposed landing site assessment methods, future changes in the target landing site are not considered. However, since the field of view and the quality of the observation data changes depending on the spacecraft state, the best landing site changes each time new observation data is obtained. This effect could not be ignored, especially when observability is limited or safe landing sites are sparse. When multiple chances of observation and divert maneuver are considered, the following two aspects should be considered. 

\begin{figure}[htbp]
    \centering
    \includegraphics[width=150mm]{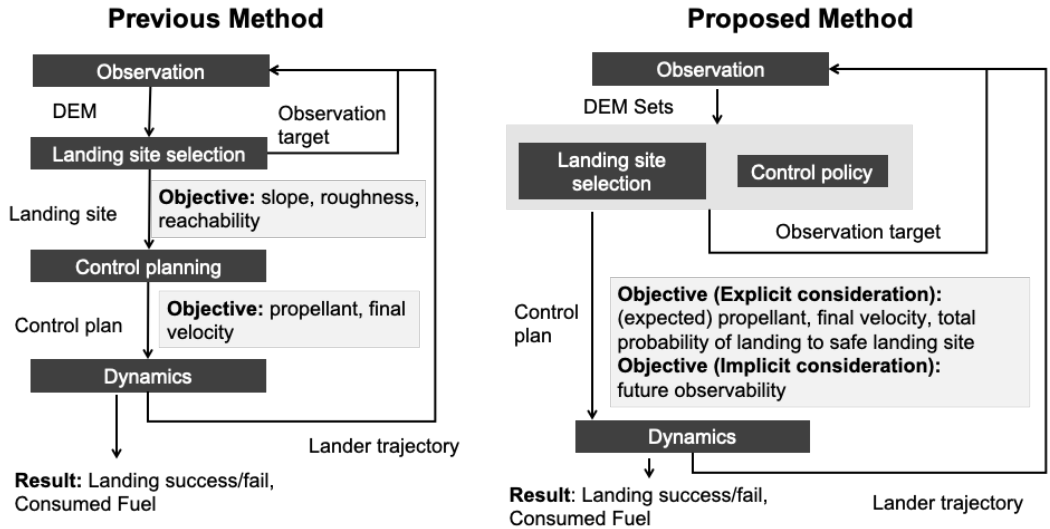}
    \caption{General Concept}
    \label{fig:genral_concept}
\end{figure}

The first aspect is the coupling between the target landing site selection and control maneuver planning. Existing landing site selection algorithms assume that the lander is guided by a simple predetermined control law, while the control law is designed to guide the lander to the decided landing target. However, when additional divert maneuvers are available in the future, the landing site selection policy is dependent on the control policy, and vice-versa.
Therefore, the landing site and control plan has to be decided simultaneously in order to achieve an overall optimal trajectory. Fig.\ref{fig:coupling_ls_control} shows an example of this coupling. Suppose landing site A and landing site B both have uncertainty in the safety level that could be judged from the observation. If there is little residual fuel, it could be safer to take the fuel optimal path to the closer landing site B (Trajectory B). However, when there is sufficient residual fuel, it could be safer to take a longer path to landing site B (Trajectory A), if additional diverting to landing site A could happen in case landing site B turns out to be unsafe as the lander approaches the ground. 

The second aspect is the observability of the terrain during descent. As illustrated in Figure \ref{fig:coupling_observability}, the change in the control policy and the target landing site affects the quality and number of the obtained observation data during descent. Therefore, the controller has to be aware that sufficient information to judge a good landing site could be obtained by following the planned trajectory. In general landing trajectory design, glide slope constraints are applied as a path constraint to ensure observations from higher elevations. However, the actual quality and quantity of observation information could not be explained only by slope angles. To tackle this problem, Crane (2013) developed an on-line information-seeking trajectory modification method that selects a trajectory that minimizes the weighted sum of the estimated entropy and field of view of the image obtained in the future, fuel consumption, and the coverage of the entire field \cite{Crane2013}. However, the research focused on modifying the nominal trajectory for a fixed landing site, and updating the landing site successively during descent was not considered in the research. The impact of landing site selection on the observation data quality should also be considered, since changing the landing site affects the observation through changes in the trajectory and the observation target, 

In order to tackle these problems, this paper proposes a learning-based method that selects landing site and design guidance trajectory successively during the HDA phase. The agent in the proposed method seeks to maximize the total probability of landing to a safe terrain landing site while minimizing total fuel consumption, landing error to the target, and final velocity. Model-free reinforcement learning techniques are leveraged to learn a policy that concurrently selects landing site and guidance strategy to the target, by interacting with the simulator environment and learning how to maximize the total probability of successful landing with minimal fuel consumption. The general concept is shown in Fig.\ref{fig:genral_concept}.

The outline is as follows. First, we explain the assumptions used in modeling the HDA phase and show that the sequence of action choices can be formulated as a partially observable Markov decision process (POMDP). Next, the method to optimize landing site selection policies and the controller is introduced. Finally, the performance of the obtained controller is analyzed, and qualitative interception is given.

\section{Problem Formulation}
\subsection{Dynamics}
A soft lunar landing scenario is assumed in this paper. In this paper, the 3 degrees of freedom (3-DOF) problem is considered. The equation of motion governing the dynamics of the problem are given as follows. 
\begin{equation}
\begin{split}
    \boldsymbol{v} &= \boldsymbol{\dot{r}}        \\
    \boldsymbol{a} &= \frac{\boldsymbol{T}}{m} + \boldsymbol{g}     \\
    \dot{m} &= -\frac{||\boldsymbol{T}||}{g_{ref} I_{sp}}   
\end{split}
\end{equation}
where $r = [r_x \ r_y \ r_z]$ is the position in the target centered orthonormal frame with the z-axis pointing upward, and $\boldsymbol{g} = [0 \ 0 \ -1.62]^T$ is the gravity. In this paper, gravity is considered to be constant during the entire mission, and the effect of planetary rotation is ignored. In addition, limitation in the thrust range is applied as follows. 
\begin{equation}
    0 \leq |\boldsymbol{T}| \leq T_{max}
\end{equation}
The following glide slope constraint are applied in most planetary landing problems
\begin{equation}
    \theta_g = \arctan{\biggl( \frac{\sqrt{r_x^2 + r_y^2}}{r_z} \biggr)}  < \theta_{gmax}
\end{equation}
This constraint is applied in order to avoid the lander from hitting the terrain at low altitudes, and to ensure that the inclination to the target is kept over
a certain amount for navigation and hazard detection purposes.

\subsection{Terrain Generation}
\begin{figure}[htbp]
 \begin{minipage}{0.45\hsize}
  \centering
   \includegraphics[width=70mm]{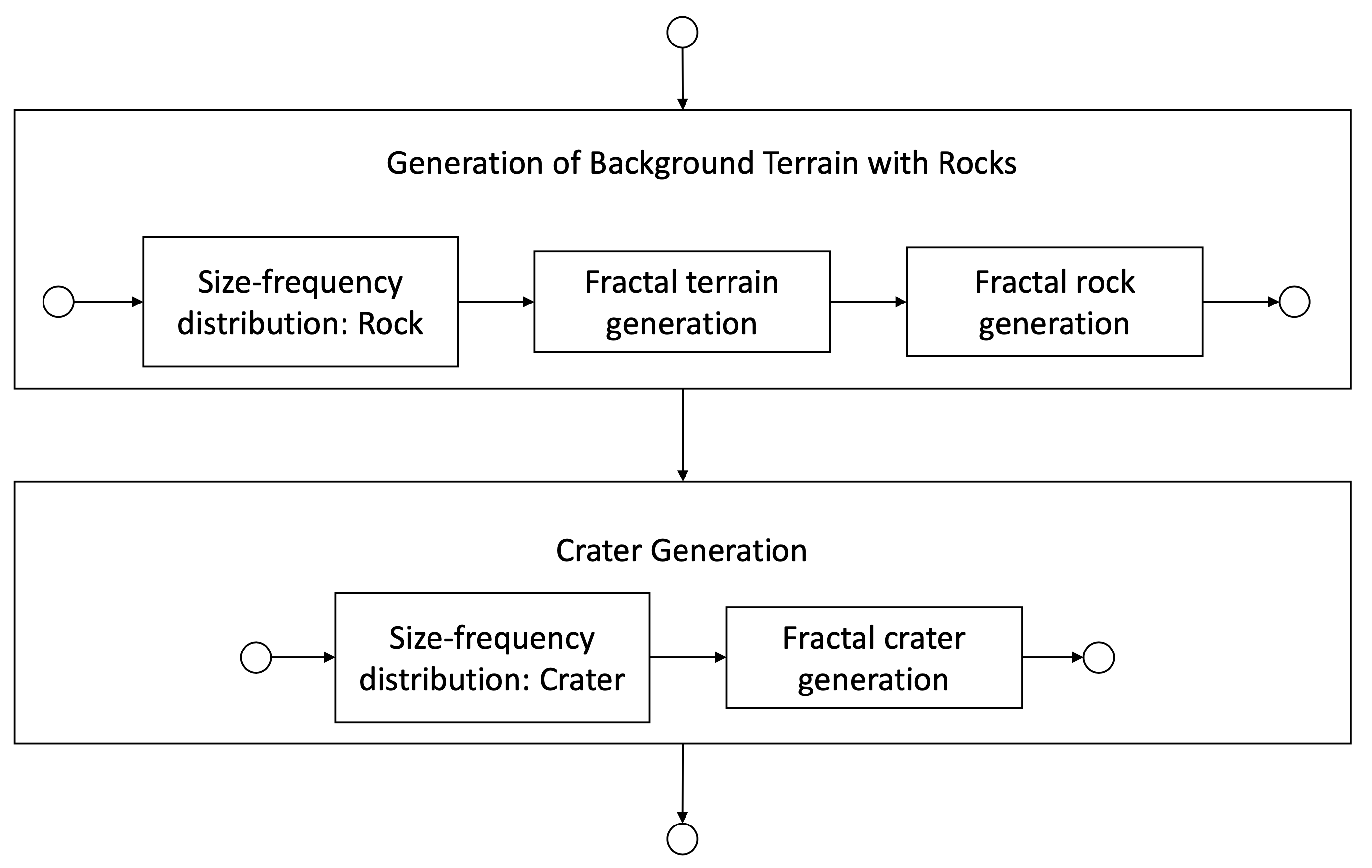}
  \caption{Terrain generation process}
  \label{fig:terrain_generation}
 \end{minipage}
 \begin{minipage}{0.45\hsize}
  \centering
   \includegraphics[width=70mm]{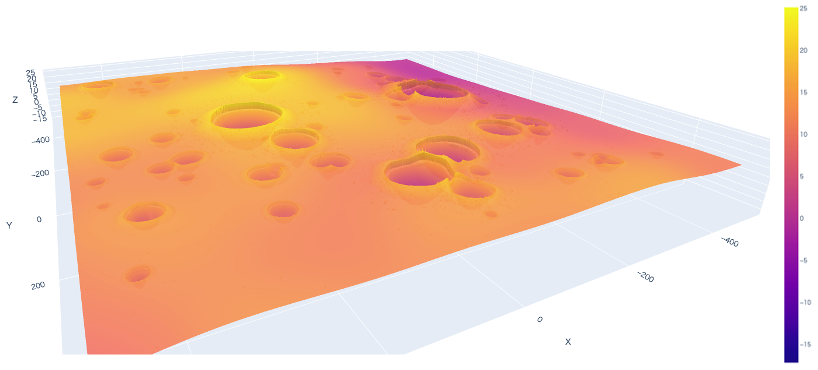}
  \caption{Example of the generated terrain}
  \label{fig:3d_terrain_map}
 \end{minipage}
\end{figure}
Observation data consists of DEM generated by HD LIDAR, and spacecraft state information (position, velocity, mass) with noise. In order to generate LIDAR DEMs, we need a true DEM as a reference. Lunar Reconnaissance Orbiter (LROC) database provides Digital Terrain Model (DTM) of the lunar surface \cite{BARKER2016346}. However, the resolution of the DTM ($\approx$ 50m/pixel) in the database is not sufficient for the LIDAR DEM model generation. Therefore, we made an original terrain generator leveraging crater distribution models proposed by Pike, and boulder distribution models proposed by Bernard \cite{Pike1977}\cite{Bernard2001}. While the model was originally developed for Mars, studies of rock density on the moon show that the model can be substituted for the lunar topography by adjusting the parameters.
The generated terrain has a size of 1000m x 1000m and a resolution of 1m. The generation process of the terrain is described in Fig.\ref{fig:terrain_generation}. An example of the generated terrain is shown in Fig.\ref{fig:3d_terrain_map}.

\subsection{Safety Map Generation}
In order to assess if the landing has succeeded in the simulator, the safety of each landing point in the generated terrain map has to be assessed. We evaluated the safety by applying the safety assessment algorithm developed in the ALHAT project to the entire generated terrain. The algorithm for assessing the deterministic safety value $V_D\in \{0,1\}$ is shown in Algorithm 1.
\begin{figure}[t!]
  \begin{algorithm}[H]
    \caption{Deterministic and Stochastic safey map generation}
    \label{alg1}
    \begin{algorithmic}
      \Require DEM $D (m \times n)$
      \Ensure deterministic safety value map $V_D (m \times n) \in \{0,1\}$, stochastic safety value map $V_P (m \times n) \in [0,1]$
      \Initialize{
      $S \leftarrow$ [0, \dots, 0]    \Comment{list of slope for each orientation}\\
      $R \leftarrow$ [0, \dots, 0]    \Comment{list of roughness for each orientation} \\
      $P \leftarrow$  [0, \dots, 0]    \Comment{list of safety probability for each orientation} \\
      $O \leftarrow$ [0, $\frac{1}{n_o}\pi, \ldots, \frac{n_o -1}{n_o}\pi$] \Comment{possible lander footpad orientation}
      }
      \For{$row=1,\cdots,m$}   \Comment{for each pixel in the DEM}
        \For{$col=1,\cdots,n$}
            \State Calculate pad placement and contact
            \For{$o_i=1, \cdots, n_o$} \Comment{calculate worst case slope for all orientations}
              \State $S[o_i] = \Call{Get\_slope}{row, col, D} $
            \EndFor
            \State $s_{max} \leftarrow \Call{Max}{S}$
            \If{$s_{max} > safe \ slope \ threshold$}  \Comment{slope not safe}
                \State $V[row][col] \leftarrow 0$
            \Else
               \For{$o_i=1, \cdots, n_o$} 
                 \State $R[o_i] = \Call{Get\_roughness}{row, col, D} $
                 \State $P[o_i] = \Call{Roughness\_To\_Safety\_Probability}{row, col, D, R} $
               \EndFor
               \State $V_P[row][col] \leftarrow \Call{Min}{P}$   
               \State $r_{max} \leftarrow \Call{Max}{R}$
               \If{$r_{max} > safe \ roughness \ threshold$}  \Comment{roughness not safe}
                  \State $V_D[row][col] \leftarrow 0$
               \Else                                       \Comment{slope and roughness safe}
                  \State $V_D[row][col] \leftarrow 1$
               \EndIf
           \EndIf
        \EndFor
      \EndFor
      \State \Return $V_D. V_P$
    \end{algorithmic}
  \end{algorithm}
\end{figure}

\subsection{Observation Data Generation}
The control agent utilizes two observation data for control: estimated lander state (position, velocity, and mass) and 2d map of the terrain. For the lander state, we assumed that we have perfect information without errors. Incorporating navigation errors and hazard relative navigation algorithms in the framework is important future work. For the 2d map of the terrain, we considered two options. The first option is to directly pass the LIDAR DEM, while the second option is to pass the stochastic safety map obtained by applying Algorithm1 on the obtained LIDAR DEM. The first option requires the control agent to assess the safety of each landing site from scratch, while in the second option, the control agent has access to more direct information about the safety of each landing point based on roughness and slope values. In this paper, we chose the second approach as we failed to design a controller using the first approach. 

The simulation of the LIDAR DEM during descent requires complex calculations. To simulate a LIDAR DEM, detector pattern formulation, ray interception with the terrain, addition of the range bias, transformation to the point clouds, and transformation to a digital elevation map has to be conducted even if we 
assume that the LIDAR position is fixed to a known position. In lunar descent cases, errors from vehicle state knowledge, LIDAR misalignment, map assembly errors when considering vehicle motions, and system latency should also have to be considered. Since the accurate modeling of the entire procedure is extremely complicated, as an initial concept study, we chose to generate pseudo observation safety map at each observation timing by cutting out the portion of the stochastic safety map of the entire field, and adding noise to it. The stochastic safety map could be obtained by directly applying Algorithm\ref{alg1} to the entire terrain DEM. By this way, we are able to greatly save the computation time required for simulation since we can avoid generating DEMs and running Algorithm\ref{alg1} at each observation timing. 

\begin{figure}[t!]
    \centering
    \includegraphics[width=80mm]{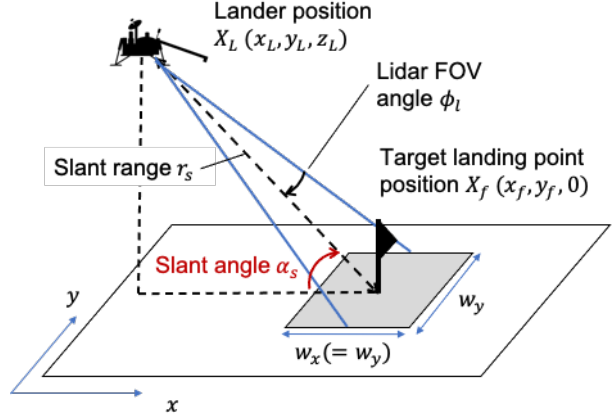}
    \caption{Illustration of the field of view of the LIDAR DEM}
    \label{fig:fovlidar}
\end{figure}

When generating the "observation safety map", three main relationship between the lander's state (or trajectory) and the LIDAR DEM (and the generated safety map) were modeled. 
The first effect is the field of view (FOV) of the LIDAR DEM. For simplicity, we assumed the obtained DEM is square, and its two axes is always aligned with the $x,y$ coordinate of the map. The length of one side of the square $w_x = w_y$ is calculated using the following equation. 
\begin{equation}
    w_x = r_s \tan{\phi_l}
\label{eq:lidarfov}
\end{equation}
where $r_s$ is the slant range (distance between the Lander and the target position), and $\phi_l$ is the field of view of the LIDAR which was set to $11.4$ [deg].(Fig. \ref{fig:fovlidar}) In reality, the DEM field of view is not rectangle, and the field of view differs between the horizontal and vertical direction of the lander, but we believe this assumption is fair enough for an initial concept study. 

The second effect is the effect of the slant range. As the slant range increases, errors in the range measurement in the LIDAR DEM increases. This effect was simulated by adding each pixel a Gaussian noise with an error proportional to the distance to each pixel. In addition, as the slant range gets longer, the sampling distance increases, small boulders in the terrain will be overlooked. This effect was modeled by fixing the LIDAR DEM size to 64x64. The third effect is the effect of slant angle. Slant angle is the angle between the horizontal plane and beam direction. As the slant angle decreases, holes in the DEM appear behind surface hazards. Therefore, regions where safety value could not be calculated appear in the safety map which leads to fewer safe regions in the map \cite{Restrepo2020}. In this paper, this effect was simply modeled by assigning unsafe labels to pixels with slant angles between the lander smaller than 70 degrees. 

In our settings, it is assumed that LIDAR DEMs could be obtained every 5 seconds, and the DEM size was fixed to 64x64. The interval time is based on the maximum processing time of the ALHAT algorithm to process DEM and generate a safety map. The entire observation data generation process is described in Algorithm 2. 

\begin{figure}[t!]
  \begin{algorithm}[H]
    \caption{Observation Data Generation at Each Timestep}
    \label{alg2}
    \begin{algorithmic}
      \Require true probabilistic safety value map of the entire terrain $V_p (1000 \times 1000)$ \\
               true position of the lander in a surface fixed frame $X_L = [x_l,y_1,z_l]$  
      \Ensure  noisy probabilistic safety value map $O_P (64 \times 64) \in [0,1]$

      \State
      \State \text{// calculate the FOV of the DEM and return 2D array of center position of each pixel in the}
      \State \text{// observed DEM}
      \State $X_D , Y_D, w_x, w_y = \Call{Calc\_FOV}{X_L, \phi_l}$   \Comment{$\phi_l$: LIDAR FOV, Eq.\ref{eq:lidarfov}}
      
       \For{$row=1,\cdots,m$}   \Comment{for each pixel in the DEM}
         \For{$col=1,\cdots,n$}
             \State $x_d, y_d = X_d[row][col], Y_d[row][col]$ 
             \State $\theta = \Call{Slant\_Angle}{x_d, y_d, X_L} $
             \If{$\theta > 20^{\circ}$}  \Comment{slant angle to the pixel is over thershold}
                \State $O_P[row][col] = 0$
             \Else      
                 \State \text{// calculate the safety probability of the pixel by finding nearest neighbor pixel in the}
                 \State \text{// whole $V_P$ map} 
                 \State $ v_p = \Call{Nearest\_Neighbor}{x_d, y_d, V_P}$ 
                 \State $ r = \Call{Slant\_Range}{x_d, y_d, X_L} $  \Comment{[m]}
                 \State $ O_P[row][col] = \Call{Clip}{v_p +  \mathcal{N}(0,\frac{r}{500} * 0.05), 0, 1} $  \Comment{Add noise to the safety value}
             \EndIf
        \EndFor
      \EndFor      
      \State \Return $O_P, w_x, w_y$
    \end{algorithmic}
  \end{algorithm}
\end{figure}

\subsection{Modeling the Problem as POMDP}
The HDA sequence is modeled as a POMDP $P = <\mathcal{S}, \mathcal{A}, T, R, \Omega, O>$ as follows. 
\begin{itemize}
    \item State space $s \in \mathcal{S}$: Spacecraft state (position, velocity, mass), true DEM of the entire field
    \item Action space $a \in \mathcal{A}$: Lander thrust output, target landing position (determines next target position for LIDAR measurement)
    \item Reward space $r \in R$: Consumed fuel, the safety of the final landing point, final velocity
    \item State transition function $T: S \times A \rightarrow \Pi(S)$: Stochastic dynamics of the spacecraft.  $T(s,a,s')$ is the probability of moving to state $s'\in \mathcal{S}$ when the agent at state $s \in \mathcal{S}$ takes action $a \in \mathcal{A}$.
    \item Observation space $\Omega$: LIDAR DEM, predicted spacecraft state
    \item Observation function $O: S \times A \rightarrow \Pi(\Omega) $: LIDAR DEM generation models, spacecraft state observation model
\end{itemize}
Obtaining an optimal policy (a policy that could maximize expected reward) in POMDP is difficult due to the partial observability of the problem. 
In general, an agent requires the entire history of the observation and action pairs $h_t = \{ a_0, r_0, o_1, \ldots,a_{t-1},r_{t-1},o_t \} $. POMDP could be converted into a belief Markov decision process (belief MDP) by introducing belief state $b$. Belief state is a conditional probability function for $s \in S$ given the history $h_t$ . When state transition function $T$ and observation function $O$ is known, an optimal policy of the POMDP could be obtained by approximating the belief MDP and applying a value iteration method.

In this paper, we seek to obtain the optimal policy of the above POMDP without modeling the $T$ and $O$ function, but by learning from transition data collected by interacting with the simulator. There are two approaches in model-free approach for POMDPs. The first approach is the memory-less approach, which learns Markov policy by simply considering the most recent observation $o$ as a state. In POMDP, the Bellman equation is not strictly satisfied, so deterministic policies leveraging only current state information is not guaranteed to be optimal. However, when the partial observability of the problem is weak, memory-less approach may be sufficient. The second approach is the memory-based approach, which learns history-dependent policy that uses the entire history data. This is usually achieved by using Recurrent Neural Networks (RNNs) that could store
history information as a single state. In this paper, both memory-less and memory-based approaches are tested.


\section{Guidance and Control}
\subsection{Observation Data Interpretation}
Since $64 \times 64$ map data is used as part of the observation, policies that take in full image data taken as an input have too large dimension to optimize. Therefore, the whole history data is transferred into a low-dimension internal state value by leveraging an auto-encoder. The role of the auto-encoder is to extract an abstract, compressed representation of the LIDAR DEM data as a latent vector $z$, to avoid directly passing down large input data to the reinforcement learning agent. The auto-encoder was trained separately with the reinforcement learning agent, as proposed in the "World Models" paper by Ha (2018). The dataset for training was collected through 5000 random rollouts, and the auto-encoder was trained to minimize the difference between the observed safety map and the reconstructed safety map produced by the decoder. The following Fig.\ref{fig:autoencoder} shows examples of the training data and the reconstructed image. 
\begin{figure}[t!]
    \centering
    \includegraphics[width=100mm]{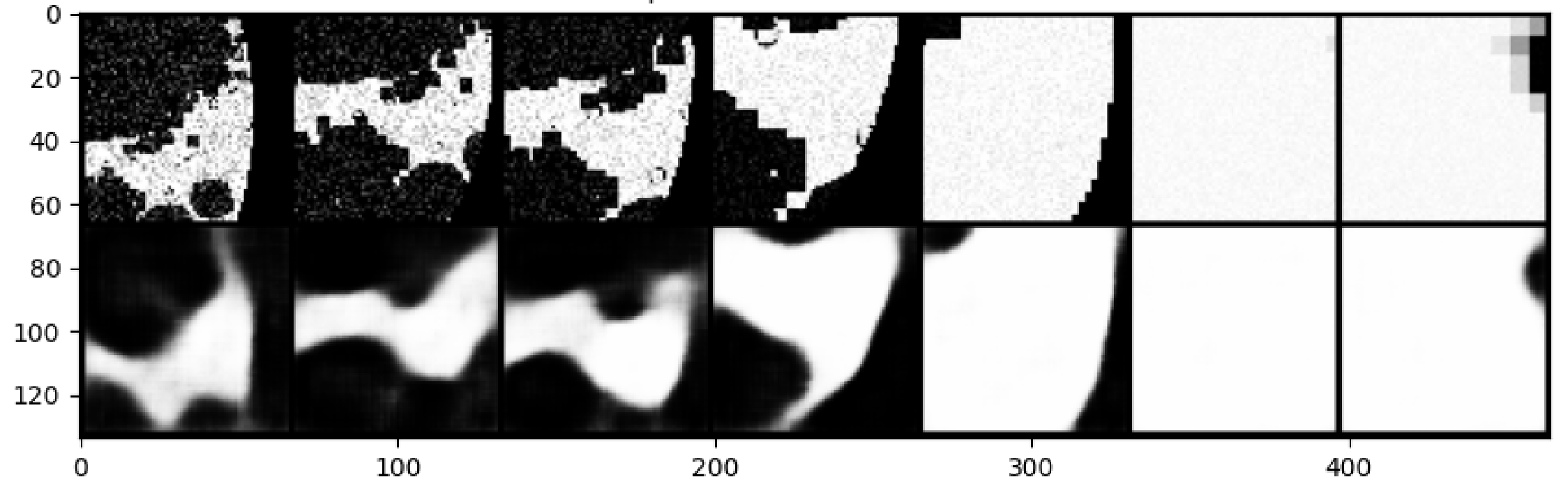}
    \caption{Original observation data (Up) and reconstructed observation data(Down) by the Autoencoder}
    \label{fig:autoencoder}
\end{figure}


\subsection{General Concept of Control}
For control, the most direct approach is to train the agent to output a three-dimensional thrust vector that guides the lander along a robust and fuel-efficient trajectory to the selected landing point. However, learning this feed-forward control policy from scratch requires high computational cost for learning. Several research have tackled this problem by shaping the reward in an effective way\cite{Gaudet2020}, but we took a different approach in order to relate the control policy with target landing points. We adopted Zero-Effort-Miss/Zero Effort-Velocity (ZEM-ZEV) feedback guidance algorithm as a baseline guidance law \cite{EBRAHIMI2008556}, and designed the agent that controls the target landing point $r_f$, and  adaptive hyper-parameters ($K_R, K_V, t_{go}$) in the ZEM-ZEV guidance algorithm. The idea of learning adaptive hyper-parameters in ZEM-ZEV feedback guidance algorithm with reinforcement learning has already been proposed in previous research, to design an ZEM-ZEV feedback controller that avoids slant angle constraint violation during descent. \cite{Furfaro2020} The difference in our paper is that the target position also changes during the descent.

\subsection{Controller}
ZEM/ZEV feedback guidance algorithm calculates the optimal acceleration using the ZEM and ZEV, which represents the distance between the final target position and velocity and the projected final position and velocity when no additional control is added from current time $t$. The optimal acceleration is calculated as 
\begin{equation}
    \boldsymbol{a} = \frac{K_R}{t_{go}^2} \boldsymbol{ZEM} - \frac{K_V}{t_{go}}\boldsymbol{ZEV}
\end{equation}
where $K_R$, $K_V$ are control gains, and $t_{go}$ is time-to-go.
When there are no limitations in the thrust magnitude and no constraints in the trajectory, it is proved that energy-optimal trajectory could be obtained by setting
the control gain as $K_R=6, K_V=2$. Energy optimal time-to-go $t_{go}$ could be obtained by solving the following equation.
\begin{equation}
    t_{go}^4\boldsymbol{g^T}\boldsymbol{g} - 2t_{go}^2(\boldsymbol{v^T}\boldsymbol{v} + \boldsymbol{v_f^T}\boldsymbol{v} + \boldsymbol{v_f^T}\boldsymbol{v_f}) + 12 t_{go} {(\boldsymbol{r_f} - \boldsymbol{r})}^T(\boldsymbol{v} + \boldsymbol{v_f}) - 18{(\boldsymbol{r_f} - \boldsymbol{r})}^T (\boldsymbol{r_f} - \boldsymbol{r}) = 0
\label{eq:optimaltof}
\end{equation}
When there is constraint $|\boldsymbol{T}| = m|\boldsymbol{a}| \leq T_{max}$ in thrust magnitude, saturated acceleration is used for control as follows.
\begin{align}
    &\boldsymbol{a}  = 
    \begin{cases}
           \boldsymbol{\bar{a}} & |\boldsymbol{\bar{a}}|\leq  T_{max}/m \\
           \boldsymbol{\bar{a}} T_{max} / m|\boldsymbol{\bar{a}}|  &  |\boldsymbol{\bar{a}}| > T_{max}/m
    \end{cases} \\
    & \boldsymbol{\bar{a}} = \frac{K_R}{t_{go}^2} \boldsymbol{ZEM} - \frac{K_V}{t_{go}}\boldsymbol{ZEV}
\end{align}
The goal of the research is to train a reinforcement learning agent that decides $K_R, K_V, t_{go}$ and target landing position $r_f$ at each time step. 

\section{Training}
\subsection{Reinforcement Learning}
In standard reinforcement learning problems, the agent interacts with the fully-observed environment $E$. At each time step $t$, the agent outputs an action $a_t$, and the environment returns the next observation $o_{t+1} = s_{t+1}$ and a step reward $r_{t+1} = g(s_t,a_t)$. The agent decides its action based on its policy $\pi(s_t)$, which maps the state to a probability distribution over possible actions. The environment follows the transition dynamics $p(s_{t+1}|s_t,a_t)$ and initial state distribution $p_{s0}$. The entire process could be modeled as a Markov decision process $\mathcal{M} = \{\mathcal{S}, \mathcal{A}, p_{s0}, p, g\}$.
The goal of the agent is to learn a policy that maximizes the expectation of discounted future reward $\mathbb{E}[R_t|\mathcal{M}, \pi]$ where $R_t = \Sigma_{i=t}^{T}\gamma^T g(s_i,a_i)$.
Reinforcement learning algorithms utilize the following action-value function (or Q-function), which describes the expected return after taking action $a_t$ in
state $s_t$, then following policy $\pi$.
\begin{equation}
    Q^{\pi}(s_t,a_t) = \mathbb{E}[R_t|s_t,a_t]
\end{equation}
For the Q function, it is known that the following recursive equation called the Bellman equation holds. 
\begin{equation}
    Q^{\pi}(s_t,a_t) = \mathbb{E}_{r_t,s_{t+1} 	\sim E}[g(s_t,a_t) + \gamma \mathbb{E}_{a_{t+1} \sim \pi}[Q^{\pi}(s_{t+1}, a_{t+1})]]
\label{eq:bellman}
\end{equation}

\subsection{Deep Deterministic Policy Gradient (DDPG)}
The Deep Deterministic Policy Gradient (DDPG) is an actor-critic, off-policy, model-free reinforcement learning algorithm for continuous action spaces.  \cite{LillicrapHPHETS15} In actor-critic methods, the critic estimates the action-value function $Q(s,a)$, while the actor produces the action given the current state based on its policy $\pi(a|s)$. As for the actor, we consider a parameterized deterministic policy $a = \pi_{\phi}(s)$ with parameter $\phi$. $\phi$ could be trained using the following deterministic policy gradient theorem and the estimated Q-values by the critic. \cite{Silver2014}
\begin{equation}
    \nabla_{\phi} J(\phi) = \mathbb{E}_{s \sim \rho_{\pi}}[\nabla_a Q(s,a|\theta)|_{s=s_t, a=\mu(s_t)} \nabla_{\phi} \mu(s|\phi)|_{s=s_t}]
\end{equation}
For the critic, in order to handle continuous state and action spaces, $Q(s,a)$ is also approximated using a function approximater parameterized by $\theta$. The critic uses the collected data to learn parameters that better approximate $Q(s,a)$. This could be achieved by minimizing the following loss $L(\theta)$
\begin{equation}
  L(\theta) = \mathbb{E}_{s_t \sim \rho^{\beta}, a_t \sim \beta, r_t \sim E}[Q(s_t,a_t|\theta) - y_t)^2]
 \label{eq:loss_critic}
\end{equation}
where $\beta$ is an arbitrary stochastic behavior policy, and $\rho^{\beta}$ is a state visitation distribution using policy $\beta$. $y_t$ is the target value
\begin{equation}
    y_t = g(s_t,a_t) + \gamma Q(s_{t+1}, a_{t+1}| \theta)
\label{eq:target}
\end{equation}
In DDPG, deterministic policy is considered. For deterministic policy $a = \mu(s)$, the inner expectation in Eq.\ref{eq:bellman} could be eliminated as follows. 
\begin{equation}
    Q^{\mu}(s_t,a_t) = \mathbb{E}_{r_t,s_{t+1} 	\sim E}[g(s_t,a_t) + \gamma Q^{\mu}(s_{t+1}, \mu(s_{t+1}))]
\end{equation}
Therefore, when calculating the loss in Eq.\ref{eq:loss_critic}, transitions obtained from different stochastic policy could be used. The collected transition data $(s_t,a_t,r_t,s_{t+1})$ is stored in a replay buffer, and then minibatch is sampled from the replay-buffer for loss calculation in Eq.\ref{eq:loss_critic}. This enables the samples for training to be distributed identically and independently which is the basic assumption for neural network training.
When exploring the environment for collecting transition data, noise is added to the actor policy. 
\begin{equation}
    \bar{\mu} (s_t) = \mu(s_t, \theta_t) + \mathcal{N}
\end{equation}
In addition, target networks are introduced in order to stabilize the training process and achieve greater convergence. When calculating Eq.\ref{eq:target} with collected transition data $(s_i,a_i,r_i,s_{i+1})$, target critic with parameter $\theta'$, and target policy with parameter $\phi'$ are used.
\begin{equation}
    y_i = r_i + \gamma Q'(s_{i+1}, \mu'(s_{i+1}| \phi')|\theta')
\end{equation}
Target networks are updated in order to slowly track the learned networks, as follows
\begin{align}
    \theta' &\leftarrow \tau \theta + (1 - \tau)\theta' \\
    \phi' &\leftarrow \tau \phi + (1 - \tau)\phi'
\end{align}
where $\tau << 1$. 

The key advantage of the DDPG method is that it enables to train deep reinforcement learning for continuous action space off-policy by assuming a deterministic policy.
This greatly improves the sample efficiency because transition data obtained by different policies in the past could be re-used for training the agent. 

\subsection{Twin Delayed Deep Deterministic Policy Gradient (TD3)}
While DDPG has achieved significant performance in continuous control problems, the critical drawback is that it was sensitive to hyper parameter settings.
This is because the overestimation of the Q-values makes the policy fall into local optima. The Twin Delayed Deep Deterministic Policy Gradient (TD3) algorithm made three major changes in the DDPG algorithm to overcome this drawback. \cite{fujimoto2018addressing}. The first change (clipped double Q-learning) is to use two seperate Q value approximators (critics), and use the smaller Q value of the two networks to calculate the target value $y_t$.
The second change is target policy smoothing. For action in the target Q value, clipped noise is added to the actual action. This prevents the Q function from having sharp peaks by returning similar Q values for similar actions. With the clipped double Q-learning and target policy smoothing, the target value from transition data $(s_i, a_i, r_i, s_{i+1})$ could be calculated as follows.
\begin{equation}
    y_i = r_i + \gamma \min_{i=1,2} Q'(s_{i+1}, \mu'(s_{i+1}| \phi') + \epsilon |\theta_i'), \quad \epsilon \sim clip(\mathcal{N}(0,\sigma),-c,c)
\end{equation}
The last change is the delayed updates of the policy and the target networks. In order to stabilize the learning process, updates of the target networks and policy networks are carried out less frequently (example: once per every $N$ steps) than the critic and networks. 

\subsection{TD3 with memory}
\begin{figure}[t!]
    \centering
    \includegraphics[width=100mm]{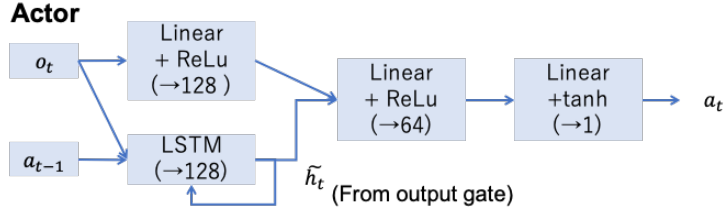}
    \caption{Architecture of the policy network}
    \label{fig:actor}
\end{figure}

\begin{figure}[t!]
    \centering
    \includegraphics[width=100mm]{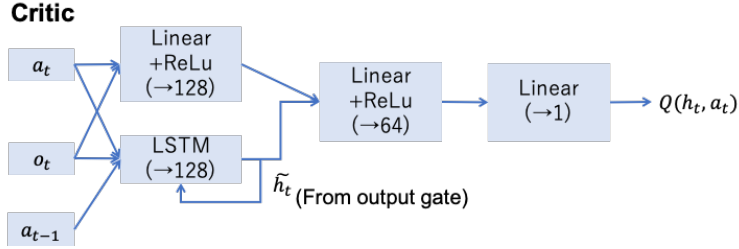}
    \caption{Architecture of the critic network}
    \label{fig:critic}
\end{figure}

When the system is partially observed, optimal policy and action-value function both become function of the entire observation-action history $h_t$. Recurrent Deterministic Policy Gradient (RDPG) algorithm uses recurrent neural networks (RNN) in the policy and the action-value networks to preserve (limited) information of the history as part of the state. \cite{Heess2015MemorybasedCW} In the RDPG setting, $\mu(s)$ and $Q(s,a)$ could be replaced with $\mu(h)$ and $Q(h,a)$. Thus, the policy update could be obtained as follows.
\begin{equation}
    \nabla_{\phi} J(\phi) = \mathbb{E}_{\tau \sim \nu_{\mu}} \biggl[ \sum_t \gamma^{t-1} \nabla_a Q(h,a|\theta)|_{h=h_t, a=\mu(h_t)} \nabla_{\phi} \mu(h|\phi)|_{h=h_t} \biggr]
\end{equation}
The critic and actor networks are where $ \tau = (s_0,o_0,a_0,s_1,o_2,a_2,\ldots) $ are drawn from the trajectory distribution $\nu_{\mu}$ generated by the current policy $\mu$ . In this paper, the RDPG algorithm was implemented by adding LSTM networks in the critic and actor networks. LSTM is a RNN architecture that is composed of a cell, an input gate, an output gate, and a forget gate. The critic and actor agent inputs the observation and action history $h_t$ to their LSTM, and uses the output of the output gate $\tilde{h_t}$ to calculate their outputs. The network structure for the critic and the actor is shown in Fig.\ref{fig:actor},\ref{fig:critic}. This architecture is based on the works of Peng (2018) of robotic control. The upper network in the figure uses the current observation (and action for critic), while the bottom network utilizes past information stored in the LSTM. 
The network was trained with the TD3 algorithm, as shown in Algorithm \ref{Alg3}. The difference from the original paper is that in our implementation, we created multiple replay buffers to store transition sequences of different episode length $T$. In our environment, the episode length varied from 4-10 time steps. We stored transition data with different length to separate replay-buffers so that we can create mini-batches with same episode length when training.

\begin{figure}[t!]
  \begin{algorithm}[H]
    \caption{TD3 with LSTM}
    \label{Alg3}
    \begin{algorithmic}
      \State Initialize Critic Network $Q_{\theta_1}, Q_{\theta_2}$, and actor network $\mu_{\phi}$  with random parameters $\theta_1', \theta_2', \mu'$
      \State Initialize target networks $\theta'_1 \leftarrow \theta_1, \theta'_2 \leftarrow \theta_2, \phi' \leftarrow \phi$
      \State Initialize list of replay buffer list $\mathcal{B}[]$
      \For{episodes = 1, M}
        \State initialize empty history $h_0$
        \State $a_{0} \leftarrow \Call{Random}{\mathcal{A}}, \quad t \leftarrow 0$
        \While{not done}
           \State $t \leftarrow t+1$
           \State receive observation $o_t$, reward $r_{t-1}$
           \State $h_t \leftarrow h_{t-1}, a_{t-1},o_t$  \Comment{Append observation and previous action to history}
           \State Select action $a_t = \mu(h_t|\phi) + \epsilon, \quad \epsilon \sim \mathcal{N}(0,\sigma)$
           \State If done $d_{t} \leftarrow 1$, else $d_{t} \leftarrow 0$
        \EndWhile
        \State Store the sequence $(a_0, o_1, r_0, d_0, \ldots, a_{t-1}, o_t, r_t, d_t)$ in $\mathcal{B}[t]$
        \State $T \leftarrow \Call{Random}{t_{max}}$   \Comment{Select the sequence length to sample from}
        \State Sample a mini-batch of $N$ episodes with length $T$:
        \State  $(a_0^i, o_1^i, r_0^i, d_1^i, \ldots, a_{T-1}^i, o_T^i, r_{T-1}^i, d_T^i)_{i=1,\ldots,N}$  from  $\mathcal{B}[T]$
        \State Construct $T \times N$ set of sequences $h_t^i = (a_0^i, o_1^i, r_0^i,d_1^i \ldots, a_{t-1}^i, o_t^i, r_{t-1}^i, d_t^i)$
        \State Compute target values for each sample episode for $t = 0,\ldots, T-1$
        \State $\quad \quad y_t^i = r_t^i + (1-d_{t+1}^i)\gamma \min_{k=1,2} Q'(h_{t+1}^i, \mu'(h_{t+1}^i| \phi') + \epsilon |\theta_k'), \quad \epsilon \sim clip(\mathcal{N}(0,\sigma),-c,c)$
        \State Update critic by minimizing the loss
        \State $\quad \quad \theta_k \leftarrow argmin_{\theta_k} \frac{1}{NT} \sum_i \sum_t (y_t^i - Q(h_t^i, a_t^i))^2$
        \State Update actor parameter $\phi$ by the deterministic policy gradient
        \State $\quad \quad \nabla_{\phi} J(\phi) =  \frac{1}{NT} \sum_i \sum_t \nabla_a Q(h,a|\theta)|_{h=h_t^i, a=\mu(h_t^i)} \nabla_{\phi} \mu(h|\phi)|_{h=h_t^i}$
      \EndFor
    \end{algorithmic}
  \end{algorithm}
\end{figure}

\subsection{Overall Framework and Training Process}
The observation $o_t$ which is used as an input of the agent is summarized in Table.\ref{tab:observations} and the output of the agent $a_t$ is summarized in Table.\ref{tab:actions}. For the training agent, we tested two architectures: a simple TD3 algorithm for memory-less approach, and a TD3 with LSTM approach for memory-based approach. For the architecture of TD3 agent without memory, we used the same architecture as the upper network of the policy and critic networks shown in Fig.\ref{fig:actor}, \ref{fig:critic}.
In order to stabilize the training, all action parameters were scaled to [0,1] during training. In the actor, the target point at timestep $t$ ($ = x_f^{(t)}, y_f^{(t)}$) is not outputted directly. Instead 2 variables $\alpha_x, \alpha_y \in [-0.25, 0.25]$ is outputted from the actor, and calculated the next target point using the following equation.
\begin{equation}
    x_f^{(t)} = x_f^{(t-1)} + \alpha_x w_x, \quad y_f^{(t)} = y_f^{(t-1)} + \alpha_y w_y
\end{equation}
In this way, the next target landing point is always selected to be within the FOV of the latest observed DEM. This prevents control failures caused by large deviations of the target between observations. The range of $\alpha_x$ and $\alpha_y$ was set to $\alpha_x, \alpha_y \in [-0.25, 0.25]$. The calculated new target point will be the center of the FOV of the DEM in the next step. Therefore, $\alpha_x$ and $\alpha_y$ are important output that has impacts on both guidance and observation. In addition, the actor does not directly output $t_{go}$ of the next step, but instead output the degrade of the time-to-go until the next observation timing which will be conducted 5 seconds later ($=\delta t_{go}$). The initial $t_{go}$ is calculated using the initial lander position and target landing site and Eq.\ref{eq:optimaltof}. 

\begin{table}[t!]
    \centering
    \caption{Elements of the observed state $o_t$}
    \begin{tabular}{l l l} \noalign{\global\arrayrulewidth=1pt} 
        \hline \noalign{\global\arrayrulewidth=0.4pt}
         content &  symbol & size  \\ \hline
         (true) lander position & $r_t$ & 3 \\
         (true) lander velocity & $v_t$ & 3 \\
         (true) lander mass     & $m_t$   & 1 \\
         width of the observed DEM range in x,y direction & $w_x^{(t)}, w_y^{(t)}$ & 2  \\
         current target position  & $x_f^{(t)}, y_f^{(t)}, z_f^{(t)}$ & 3  \\
         latent vector of the generated safety map encoded through autoencoder & $z_t$ & 32 \\ \hline
         total & & 44 \\ 
        \noalign{\global\arrayrulewidth=1pt} \hline \noalign{\global\arrayrulewidth=0.4pt}
    \end{tabular}
    \label{tab:observations}
\end{table}

\begin{table}[t!]
    \centering
    \caption{Elements of the action $a_t$}
    \begin{tabular}{l l l l} 
         \noalign{\global\arrayrulewidth=1pt} \hline \noalign{\global\arrayrulewidth=0.4pt}
         content &  symbol & size  & range\\ \hline
         gain of the ZEM-ZEV control & $K_R$ & 1 & [5, 7] \\
         gain of the ZEM-ZEV control & $K_V$ & 1 & [1, 3] \\
         degradation of time-to-go from previous timestep & $\delta t_{go}$ & 1 & [4.25,5.75] \\
         target landing point position within the DEM FOV& $\alpha_x, \alpha_y$  & 2  & [-0.5, 0.5]\\ \hline
         total & & 5 \\ 
        \noalign{\global\arrayrulewidth=1pt} \hline \noalign{\global\arrayrulewidth=0.4pt}
    \end{tabular}
    \label{tab:actions}
\end{table}

\begin{figure}[t!]
    \centering
    \includegraphics[width=\hsize]{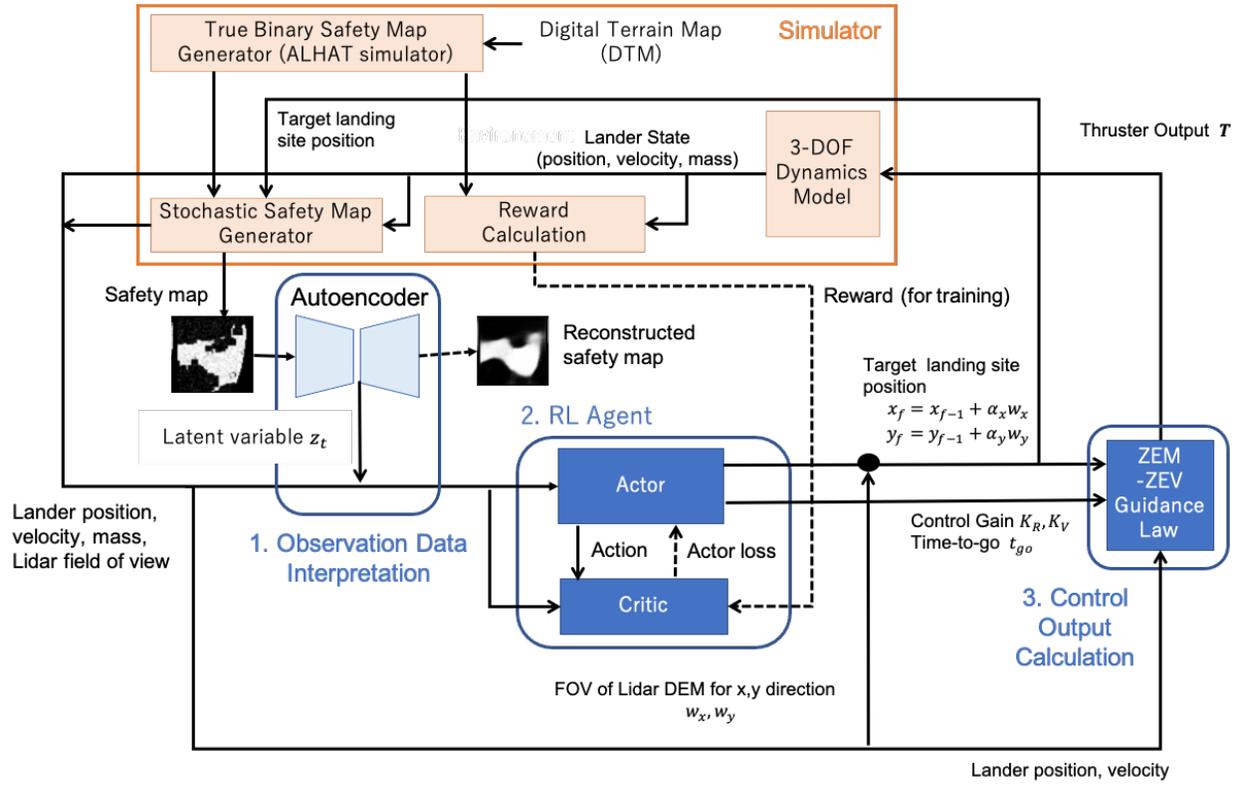}
    \caption{Overall framework}
    \label{fig:framework}
\end{figure}

Reward setting is the key in reinforcement learning. The reward at time step $t$ was implemented as follows.
\begin{align}
\begin{aligned}
    r_t &= \alpha_m \frac{m_{t-1} - m_t}{m_0 - m_{dry}} + \alpha_f (U(z_{max} - z_t) + U(m_{dry} - m_t)) + \alpha_v d_t (|\boldsymbol{v_t}|) \\
           & + \alpha_r d_t (|\boldsymbol{r_t} - \boldsymbol{r_f^{(t-1)}}|) + \alpha_s d_t V_D(r_t)
\end{aligned}
\end{align}
where
\begin{align}
    \begin{aligned}
       U(x) &= \begin{cases}
          1 & \text{if $x\geq 0$} \\
          0 & \text{otherwise}
       \end{cases} \\ 
       d_t &= \begin{cases}
          1 & \text{if episode done} \\
          0 & \text{otherwise}
       \end{cases} \\ 
       V_D(r_t) &= \begin{cases}
          1 & \text{if $r_t$ is a safe landing point} \\
          -1 & \text{if $r_t$ is not a safe landing point}
       \end{cases} \\ 
       z_{max} &= 50 \quad \text{(maximum height of the terrain)}
    \end{aligned}
\end{align}
The term with $\alpha_m$ is a penalty for fuel consumption. The term with $\alpha_f$ is a penalty term for breaking the two critical path constraints: hitting the ground and running out of fuel. The term with $\alpha_v$ is a penalty for velocity norm at final episode to ensure soft landing, and  
the term with $\alpha_r$ is a penalty for not landing to the final target landing point. These two terms are required because ZEM-ZEV control with adaptive gain and $t_{go}$ parameters are not guaranteed to achieve a pinpoint soft landing to the target. Setting the weights $\alpha_m, \alpha_v, \alpha_r, \alpha_s$ is a difficult problem which depends not only on the mission designer focus but the stability of the training process. It is natural to assume that $\alpha_f, \alpha_v$ and $\alpha_s$ should take relatively large values because hitting the ground at a high rate of speed and landing in a danger zone could both lead to the immediate loss of the lander. We used $\alpha_m=1, \alpha_f=-10, \alpha_v=0.1, \alpha_r=0.01, \alpha_s=1$ in our research. Note that slope angle constraints are not considered as penalty in the reward settings. This is because slope angle constraints are not directly related to the primary objective of the planetary landing, but rather conservative constraints to assure preferable geometry for observation. By interacting with the simulator that simulates the observation process and the dynamics, we expect the agent to learn to take a trajectory with adequate slope angle in order to achieve safe landing. The overall framework is summarized in Fig \ref{fig:framework}.

\section{Simulation}
\subsection{Initial Condition and Hyper-parameter Settings}
The initial state and other specs of the lander are summarized in Table.\ref{tab:conditions}. Parameters that have range are randomly initialized within the range every time the episode starts. The true DTM is picked randomly from the list of DTMs generated before the training.

\begin{table}[t!]
    \centering
    \caption{Simulation condition}
    \begin{tabular}{l l l} 
         \noalign{\global\arrayrulewidth=1pt} \hline \noalign{\global\arrayrulewidth=0.4pt}
         content &  symbol & value \\ \hline
         Initial altitude (downrange) [m] & $z_0$ & [900, 1000]  \\
         Initial crossrange distance [m] & $\sqrt{x_0^2 + y_0^2} $ & [200 250] \\
         Initial downrange velocity [m/s]& $|v_z| $ & [20, 35] \\
         Initial crossrange velocity [m/s] & $|v_z|$ & [5, 10] \\
         Initial mass of the lander [kg] & $m_0$ & 1200 \\
         Altitude of the final target point [m] &$z_f$ & 50 \\
         Dry mass of the lander [kg] & $m_{dry}$ & 1150 \\
         Specific Impulse of the lander [s] & $I_{sp}$ & 325 \\
         Maximum thrust of the lander [N] & $T_{max}$ & 12000 \\
         Thrust magnitude error ratio & $ |T - T_{plan}|/T_{plan}$ & 0.05 \\
         Number of different DTMs used for training & $N_{D-training}$  & 70  \\ 
         Number of different DTMs used for variation & $N_{D-variation}$  & 18  \\ 
        \noalign{\global\arrayrulewidth=1pt} \hline \noalign{\global\arrayrulewidth=0.4pt}
    \end{tabular}
    \label{tab:conditions}
\end{table}

\subsection{Training Process}
\begin{figure}[t!]
    \centering
    \includegraphics[width=150mm]{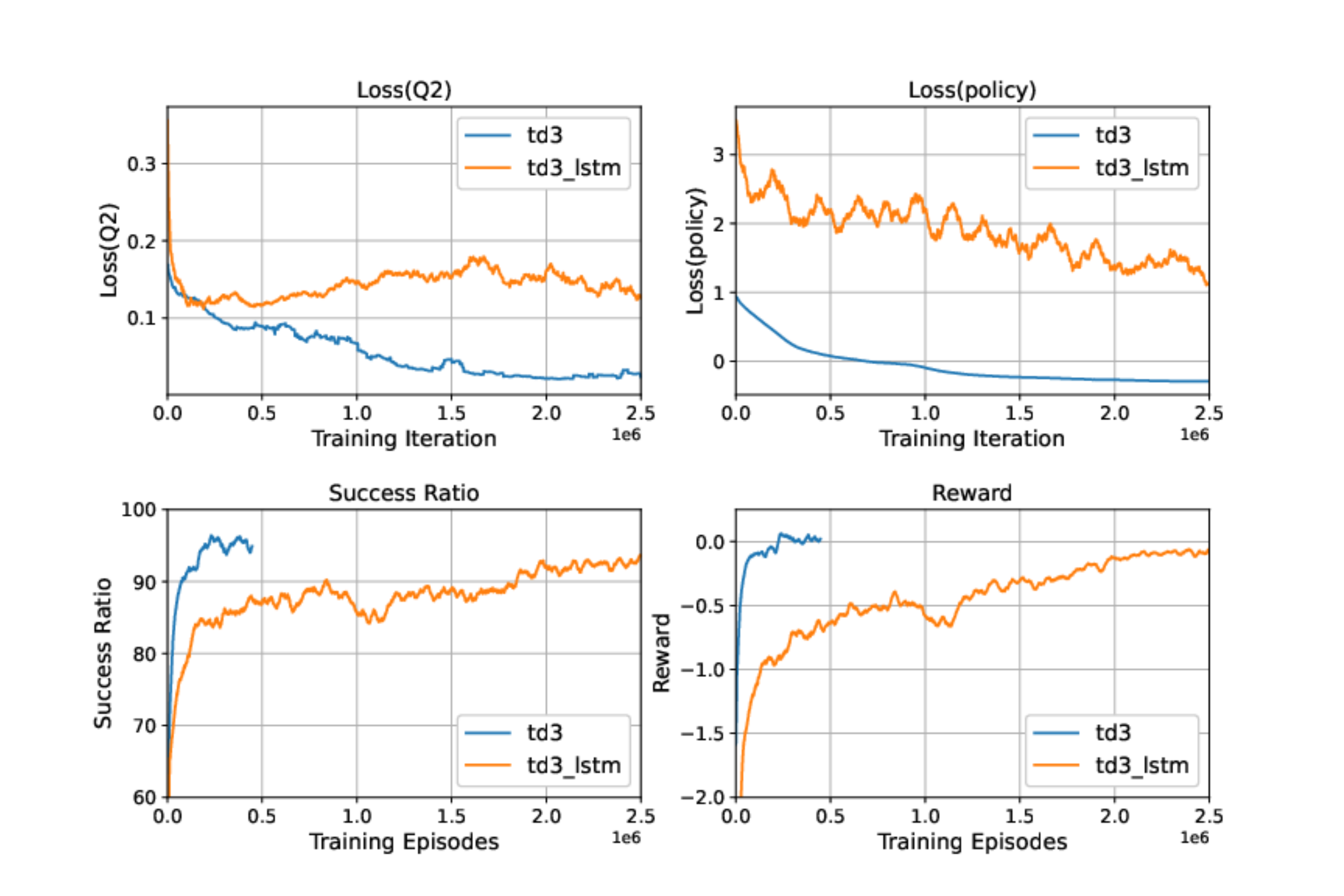}
    \caption{Training log. It should be noted that the x-axis of the above two figures represent the number of training updates, while the x-axis of the bottom two figures represent the number of training episodes.}
    \label{fig:training_log}
\end{figure}

Fig.\ref{fig:training_log} illustrates the transition of the training loss of the critic and actor network, reward, and the ratio of landing to safe landing site during training, for both the agent without LSTM and with LSTM. Values in the graph represent the moving average of 5000 episodes. When validating the performance during training, DTM datasets that are different from training datasets were used. It required around 250k iteration of training for both agents to reach around the maximum performance. 

Contrary to our expectations, the agent without LSTM showed better performance compared with the agent with LSTM. The training of agent with LSTM was much more sensitive to its hyper-parameters, and we did not manage to achieve higher performance than TD3 without memory. Another reason why agent without LSTM could achieve high performance is likely because the partial observability of our problem is not strong, thanks to the access to the stochastic safety map with relatively low errors. Adding larger errors into the observation data such as navigation errors might require the use of memory to cope with strong partial observability.

\subsection{Comparison with other strategies}
In order to assess the performance of the trained agent, the performance of the two agents were compared with two other strategies. 
The first strategy is the "Fixed Control Policy". This policy ignores the $K_R, K_V, t_{go}$ output from the agent, and uses the agent only to select the target point. The gains are fixed to $K_R=6, K_V=2$, and $t_{go}$ was calculated by Eq.\ref{eq:optimaltof}. This policy was introduced to see how the agent's adjustment of gain and flight time affected the trajectory, final velocity, and landing errors to the target. The second strategy is the "Single Divert Policy". This strategy makes no use of trained agents. This policy targets the initial target point until the altitude reaches 500m, and then selects the pixel with maximum safety value in the stochastic safety map obtained at 500m altitude as the final target point. The gains are fixed to $K_R=6, K_V=2$, and $t_{go}$ was calculated by Eq.\ref{eq:optimaltof}.
This policy was introduced to represent a simplified version of existing landing techniques. 

\begin{figure}[t!]
    \centering
    \includegraphics[width=\hsize]{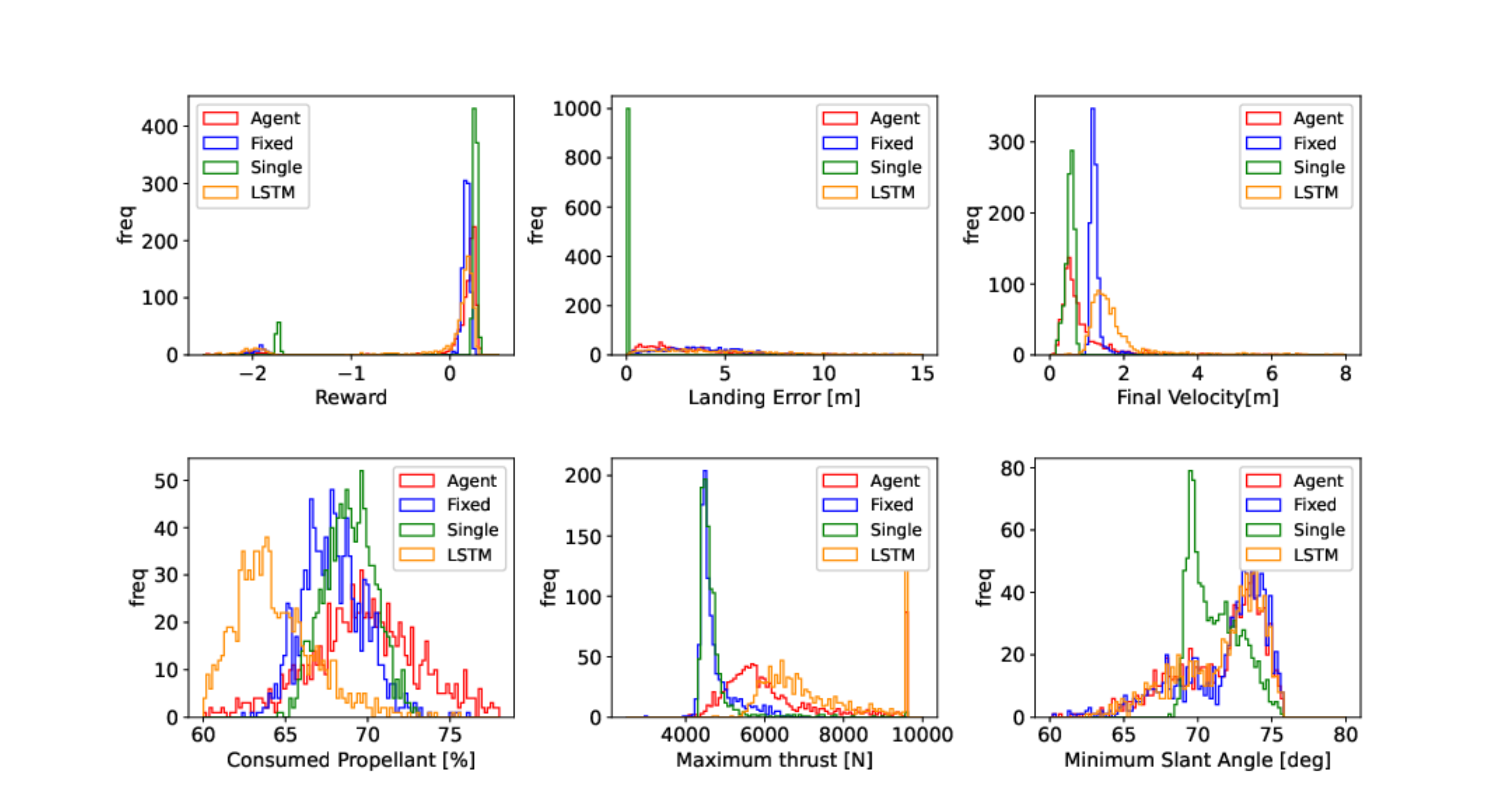}
    \caption{Evaluation of performance. 'Agent' is the policy of the trained memory-less agent, 'LSTM' is the policy of the trained memory-based agent, 'Fixed' is the policy with fixed ZEM-ZEV parameters, and 'Single' is the policy that conducts single divert maneuver and selects the landing site by directly using the stochastic safety map.}
    \label{fig:histogram}
\end{figure}

The performance of the four policies was tested by landing simulation of 1000 episodes with random initial conditions. In order to assess the performance in cases that the agent needs to change its initial landing site, the initial target position was chosen so that at least 80\% of the area within 100 meter radius is hazardous. 
Fig.\ref{fig:histogram} shows the histogram of the reward, landing error to the final target point[m], final velocity magnitude[m/s], maximum thrust[N], and minimum slant angle [deg]. The mean value is summarized in Table.\ref{tab:comparison}.

\begin{figure}[t!]
\begin{subfigmatrix}{2}
  \subfigure[Trajectory by agent controller (White:Safe Black:Unsafe) ]{\includegraphics[height=80mm]{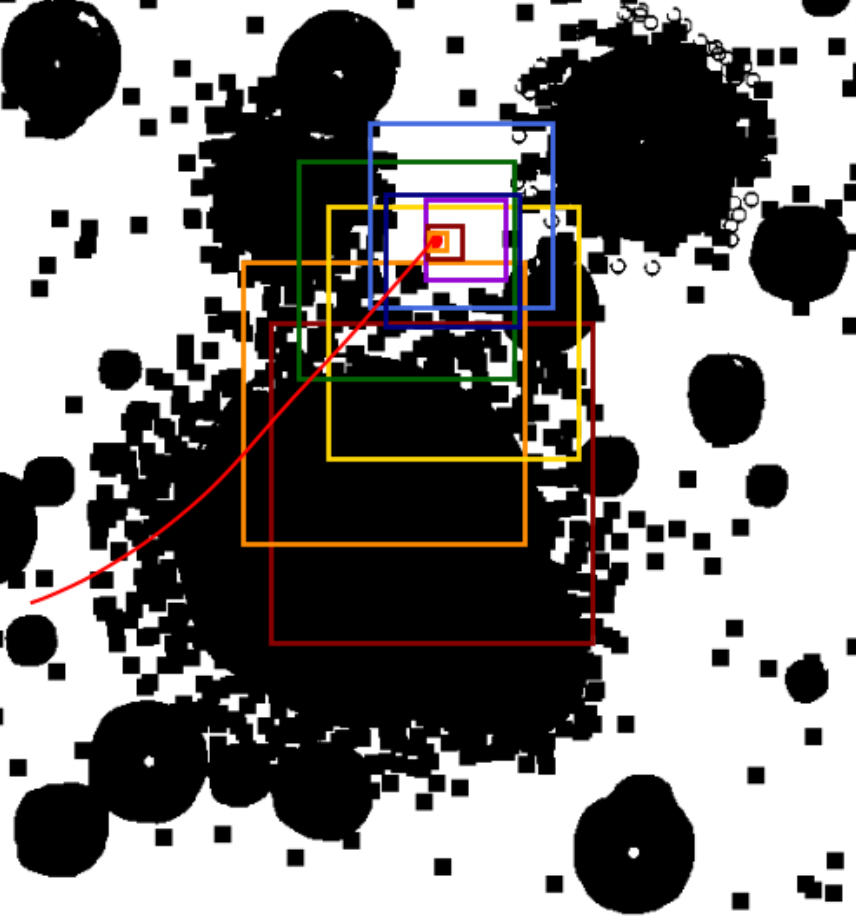}}
  \subfigure[Control history by the agent controller (Up:Velocity Down:Thrust)]{ \includegraphics[height=80mm]{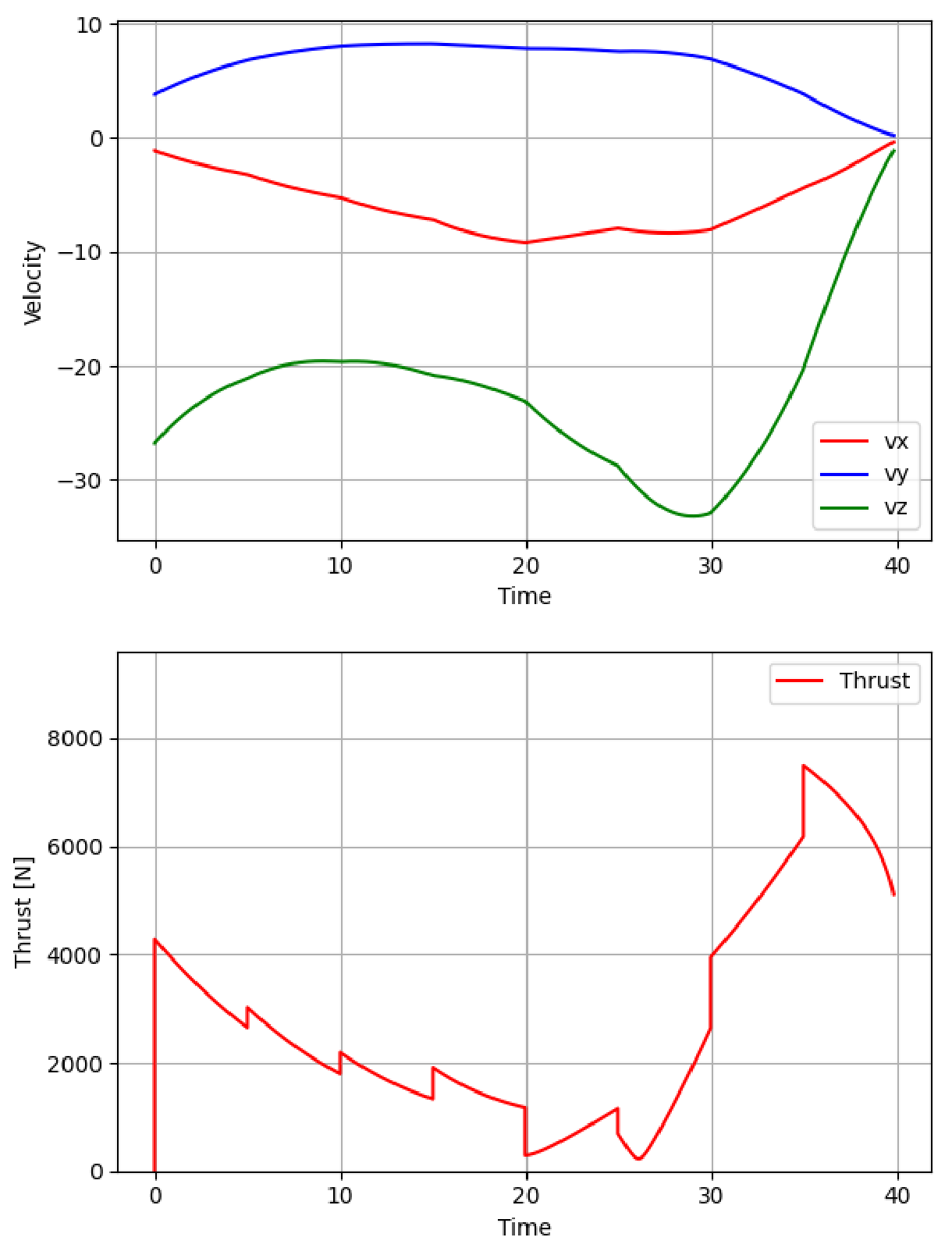}}
   \subfigure[Trajectory by fixed controller  (White:Safe Black:Unsafe)]{\includegraphics[height=80mm]{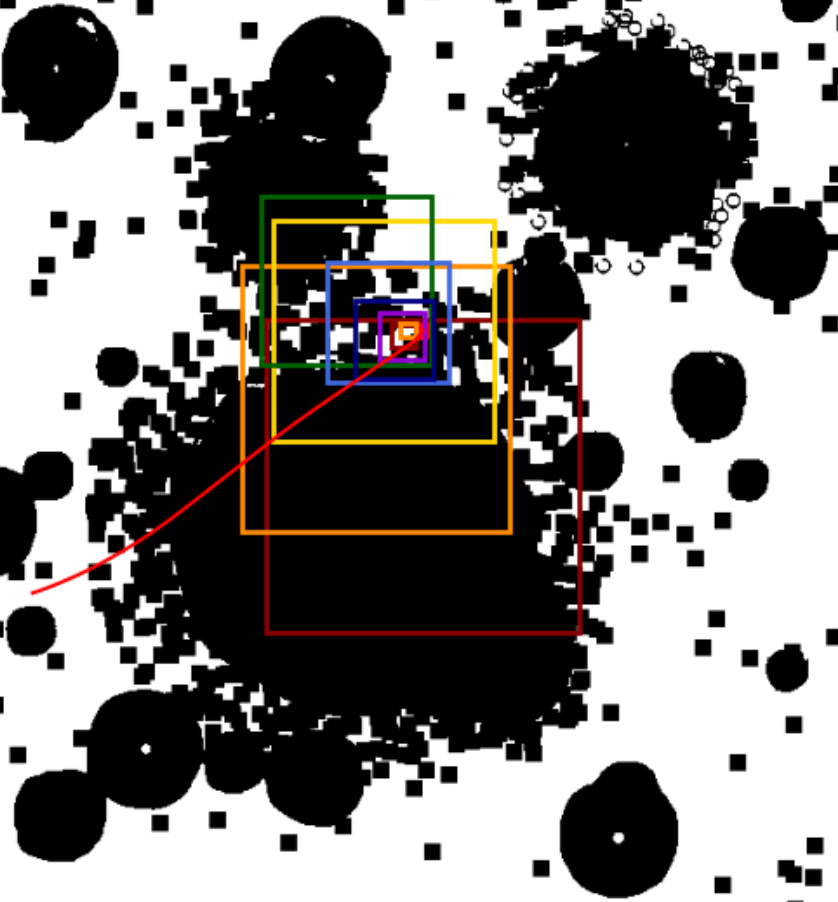}}
  \subfigure[Control history by the fixed controller ((Up:Velocity Down:Thrust)]{ \includegraphics[height=80mm]{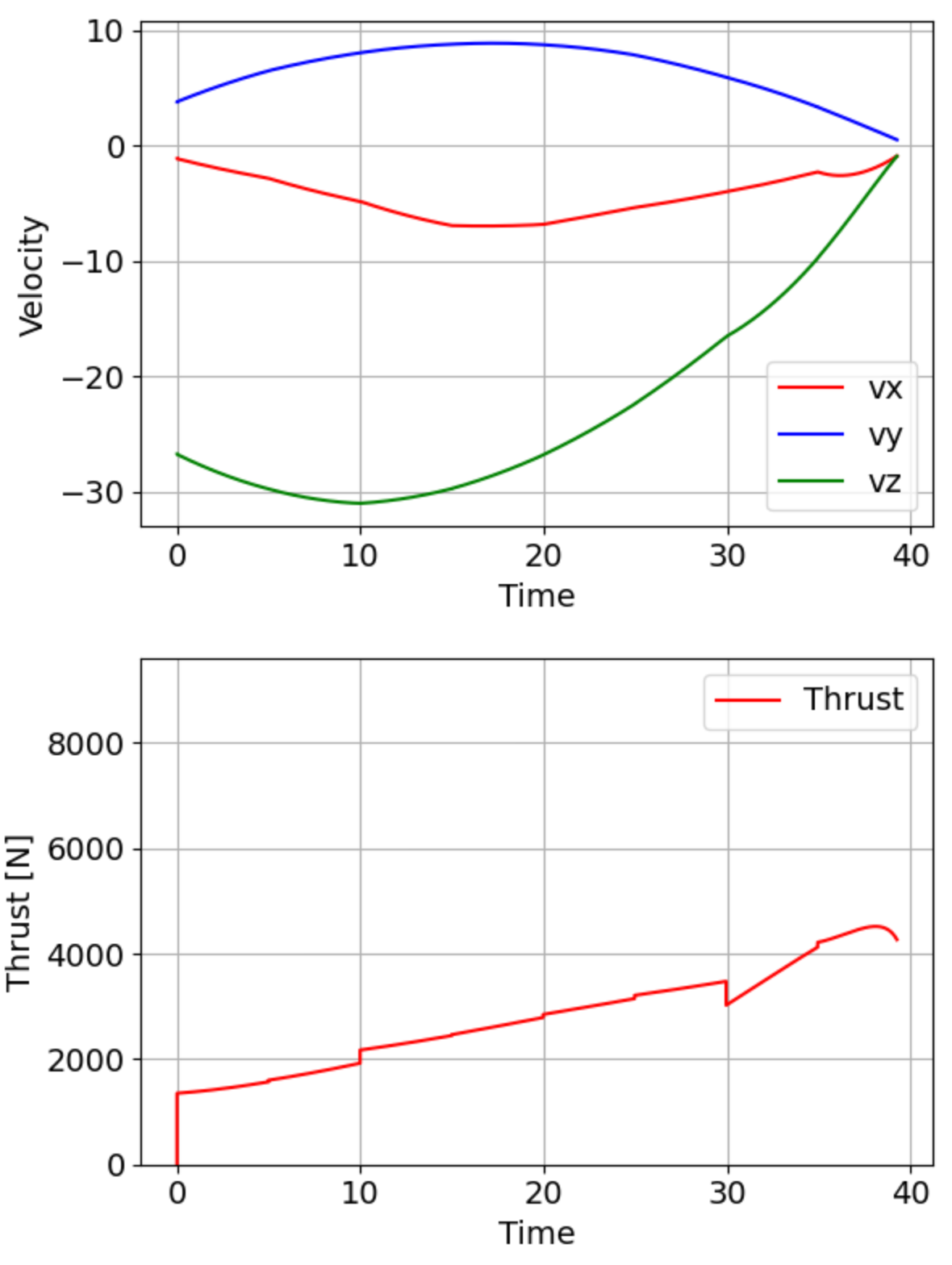}}
\end{subfigmatrix}
\caption{Comparison of the obtained agent and a ZEM-ZEV controller with fixed gain. Red line in figure(a),(c) shows the lander trajectory, while the square shows the
field of view (FOV) of the LIDAR DEMs.}
\label{fig:comparison}
\end{figure}

The trained memory-less agent achieved maximum ratio ($94.8\%$) of landing to safe landing site. 
The fixed control policy had a lower safe landing ratio to the agent policy, due to its limitation in changing the thrust magnitude flexibly when long divert maneuver is required. Fig.\ref{fig:comparison} shows an example of such a situation.
In this case, the initial target point (the center of the red square) is inside the hazardous region (black area of the figure), and the target landing point has to be shifted toward upside of the figure in order to achieve a safe landing. While the trained agent reduces the $z$ axis velocity immediately as shown in Fig.\ref{fig:comparison}-(b) in order to maintain the altitude and wide field of view for searching safe landing sites, the fixed controller does not aggressively thrust the propellant as shown in Fig.\ref{fig:comparison}-(d). Therefore the field of view of the fixed gain controller is kept limited, and the fixed controller fails in finding a wide safe area to land. 

The trained agent with memory had the lowest probability of landing to safe landing site, and the largest landing position error and velocity error among the four policies. On the contrary, the average fuel consumption was the lowest among the four policies. This implies that the agent was trapped in a local optima policy that increases reward by reducing fuel consumption rather than improving other metrics.

The single observation policy fails in selecting safe landing sites when safe landing sites could not be obtained in the single observation due to limitations in the field of view or slant angles. The rate of failure may be reduced by selecting optimal altitude for the divert decision or implementing better landing site selection algorithms from safety maps. However, it should be noted that in reality, the risk of single observation policy is likely to increase than our simulator due to additional error sources (navigation errors, attitude constraints, etc) which are not simulated in our research. These errors degrade the quality of the DEM (or safety map) and the accuracy of lander guidance to the target point. 

\begin{table}[t!]
    \centering
    \caption{Comparison of performance between methods (1000 episodes. Mean value)}
    \begin{tabular}{l c c c c} 
         \noalign{\global\arrayrulewidth=1pt} \hline \noalign{\global\arrayrulewidth=0.4pt}
         Criteria &  Memory-Less Agent &  Memory-Based & Fixed Control  & Single Divert \\ \hline
         Safe Landing Ratio [\%] & 94.8 \% & 88.3 \%  & 93.2 \% & 89.7 \%  \\
         Distance to final target [m] & 2.492  & 4.547  & 3.786  & 0.005 \\
         Final Velocity [m/s] & 0.637 & 1.520 & 1.189  & 0.557 \\
         Propellant Consumption [\%] & 69.88 & 63.70 & 67.84 & 68.89 \\
         Minimum Slant Angle [deg] & 72.429 & 72.570 & 72.795  & 70.85 \\
         Maximum Thrust [N] & 5818 & 6863 & 4553 & 4556 \\
        \noalign{\global\arrayrulewidth=1pt} \hline \noalign{\global\arrayrulewidth=0.4pt}
    \end{tabular}
    \label{tab:comparison}
\end{table}

\begin{figure}[h]
    \centering
    \includegraphics[width=65mm]{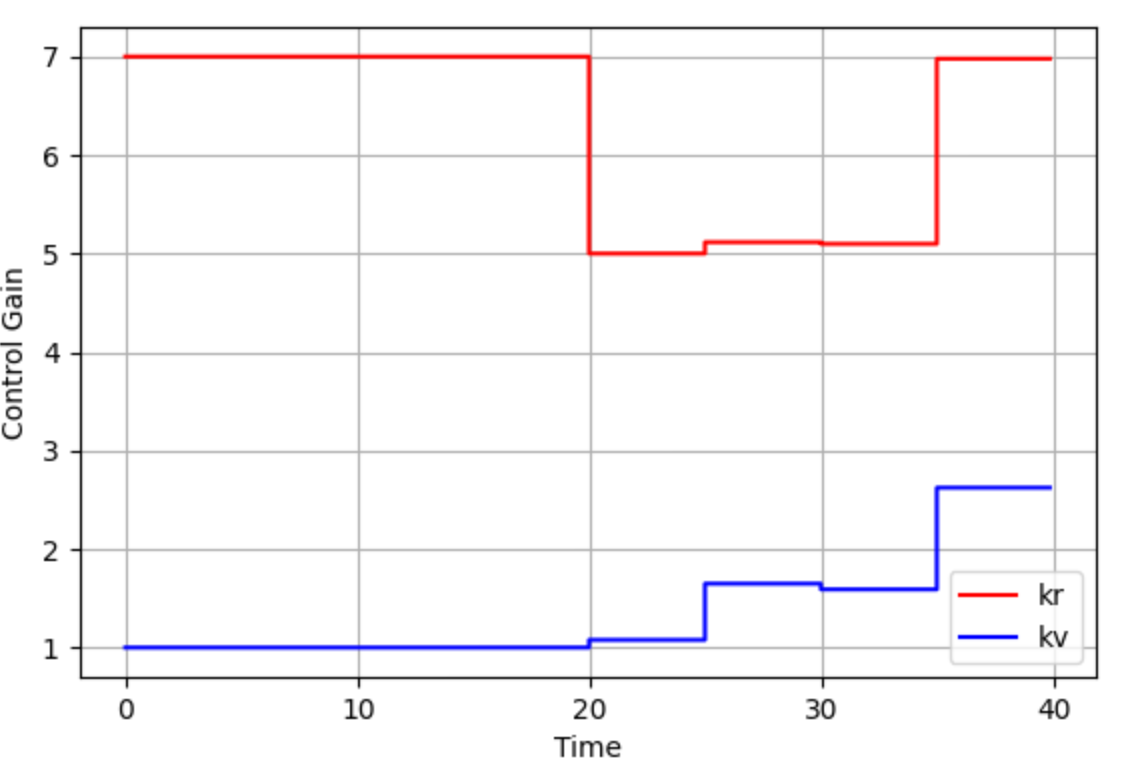}
    \caption{Changes in the control gain $K_R, K_V$ while control in Fig.\ref{fig:comparison}-(a),(b)}
    \label{fig:my_label}
\end{figure}

While the trained agent showed high potential in adaptly changing its target position and control outputs to find safe landing sites, they also have limitations in reducing velocities and landing error to the final target point decided by the agent. Since convergence to the target position and velocity is not guaranteed when control gain in ZEM-ZEV control is changed, this disadvantage is inevitable without further refinements. In addition, fuel consumption of the memory-less trained agent was larger compared with the two agents for comparison, due to the large thrust magnitude during descent and frequent changes of the target point. Addressing the trade-off between safety and fuel consumption from the point of the frequency of the target point changes is an interesting future work. 




\section{Conclusion}
In this paper, we proposed a new learning-based framework for Hazard Detection and Avoidance (HDA) phase which successively updates the target landing site and control parameters simultaneously after each observation, in order to cope with the coupling between observation, guidance, and control. We modeled the HDA sequence as a POMDP, and a reinforcement learning agent that interprets the obtained map using auto-encoder and outputs control parameters for ZEM-ZEV feedback control law was developed to find an optimal policy. The agent was trained by interacting with the simulator, and the trained agent was able to achieve over 90\% probability of successful landing at difficult landing sites where over 80\% of the terrain was hazardous around the initial target point, by gradually updating the target landing point towards safe regions. 

In order to incorporate a more realistic and higher level of uncertainty to the environment, our future work will mainly focus on the refinement of the simulator model. This includes the incorporation of accurate LIDAR DEM generation models, expansion of dynamics to 6-DOF, and incorporation of lander state measurement errors. As the partial observability of the environment increases by incorporating various error sources, a memory-based agent might be required for sufficient performance. We are also planning to test other baseline control policies and observation data interpretation architectures to improve the agent's performance and stability. 

\section{Acknowledgements}
This material is partially based upon work supported by the National Aeronautics and Space Administration under Grant No.80NSSC20K0064 through the NASA Early Career Faculty Program.

\newpage
\bibliographystyle{AAS_publication}   
\bibliography{references}   

\end{document}